\pdfoutput=1
\documentclass[11pt]{article}

\usepackage[preprint]{acl}

\usepackage{times}
\usepackage{latexsym}

\usepackage[T1]{fontenc}

\usepackage[utf8]{inputenc}

\usepackage{microtype}

\usepackage{inconsolata}

\usepackage{amsmath}
\usepackage{amssymb}
\usepackage{graphicx}
\usepackage{booktabs}
\usepackage{multirow}
\usepackage[table]{xcolor}
\usepackage{arydshln}
\usepackage[most]{tcolorbox}
\usepackage{caption}
\usepackage{subcaption}
\usepackage{pifont}
\newcommand{\cmark}{\textcolor{green!60!black}{\ding{51}}}
\newcommand{\xmark}{\textcolor{red!70!black}{\ding{55}}}

\definecolor{sysblue}{HTML}{E8F0FE}
\definecolor{sysborder}{HTML}{4285F4}
\definecolor{usrgreen}{HTML}{E6F4EA}
\definecolor{usrborder}{HTML}{34A853}

\newtcolorbox{promptbox}{
  enhanced,
  width=\textwidth,
  colback=white, colframe=white,
  boxrule=0pt, arc=0pt,
  top=0pt, bottom=0pt, left=0pt, right=0pt,
}

\newtcolorbox{sysmsg}{
  enhanced, nobeforeafter,
  colback=sysblue, colframe=sysborder,
  boxrule=0.3pt, arc=2pt,
  left=5pt, right=5pt, top=3pt, bottom=3pt,
  before upper={\textbf{\sffamily\footnotesize System}\par\smallskip\ttfamily\tiny},
}

\newtcolorbox{usrmsg}{
  enhanced, nobeforeafter,
  colback=usrgreen, colframe=usrborder,
  boxrule=0.3pt, arc=2pt,
  left=5pt, right=5pt, top=3pt, bottom=3pt,
  before upper={\textbf{\sffamily\footnotesize User}\par\smallskip\ttfamily\tiny},
}

\title{Can LLMs Access their World Knowledge for Event Prediction?}

\author{
\textbf{Zijun Wu}\textsuperscript{1,2}\thanks{Work done during an internship at RBC Borealis.},
\textbf{Zhitian Zhang}\textsuperscript{1},
\textbf{Hossein Hajimirsadeghi}\textsuperscript{1} \\
\textbf{Nikita Dvornik}\textsuperscript{1},
\textbf{Alexander Pashevich}\textsuperscript{1} \\
\textsuperscript{1}RBC Borealis
\quad
\textsuperscript{2}University of Alberta \\
\texttt{zijun4@ualberta.ca} \\
\texttt{\{andy.zhang,nikita.dvornik\}@rbc.com} \\
\texttt{\{hossein.hajimirsadeghi,alexander.pashevich\}@borealisai.com}
}

\begin{document}
\maketitle

\begin{abstract}

Recent work has questioned whether large language models (LLMs) can be useful for temporal event prediction. Despite vast pretrained world knowledge, LLMs often underperform simpler specialized models. We ask whether this knowledge is truly inaccessible, or whether current methods fail to present event prediction in a form LLMs can use. We argue for the latter and show that LLMs become effective event forecasters when the interface is designed around two choices. The first grounds each event in a natural-language description, and the second separates event semantics from the prediction space to prevent semantic bias from label associations. To test this, we introduce \textsc{DescribeEvent}, a benchmark of $\sim$150K events across seven datasets and six domains with event-level descriptions. We find that capable LLMs substantially benefit from descriptions and symbolic labels, outperforming description-free variants and non-LLM baselines, while weaker models gain little.
These results suggest that accessing LLM world knowledge for event prediction is an interface problem, and that event descriptions offer a practical way to elicit the knowledge LLMs already encode.\footnote{Our code and dataset are available at \url{https://github.com/BorealisAI/DescribeEvent}.}

\end{abstract}

\section{Introduction}

Temporal event prediction studies how events evolve over time and asks what event will happen next, and when. This problem is typically modeled with temporal point processes (TPPs), which represent each event as a categorical type paired with a timestamp~\cite{hawkes1971spectra,du2016recurrent,mei2017neural}.
Recent work are adapting LLMs to temporal event prediction, motivated by the assumption that their pretrained world knowledge can help infer the next event~\cite{liu2024tppllm,kong2026languagetpp,chen2025tpptal,shi2023lamp,gruver2023llmtime,zhou2023onefitsall,requeima2024llmprocesses, gupta2025laststop}. However, recent evidence challenges this view: LLM components can often be removed with little loss, and LLM-based forecasters often fail to beat simpler specialized models~\cite{tan2024lmts, Chang2024ACE}. This raises our central question: \emph{is LLM knowledge inaccessible for event prediction, or have current methods failed to make it usable?}

\begin{figure*}[t]
  \centering
  \includegraphics[width=\textwidth]{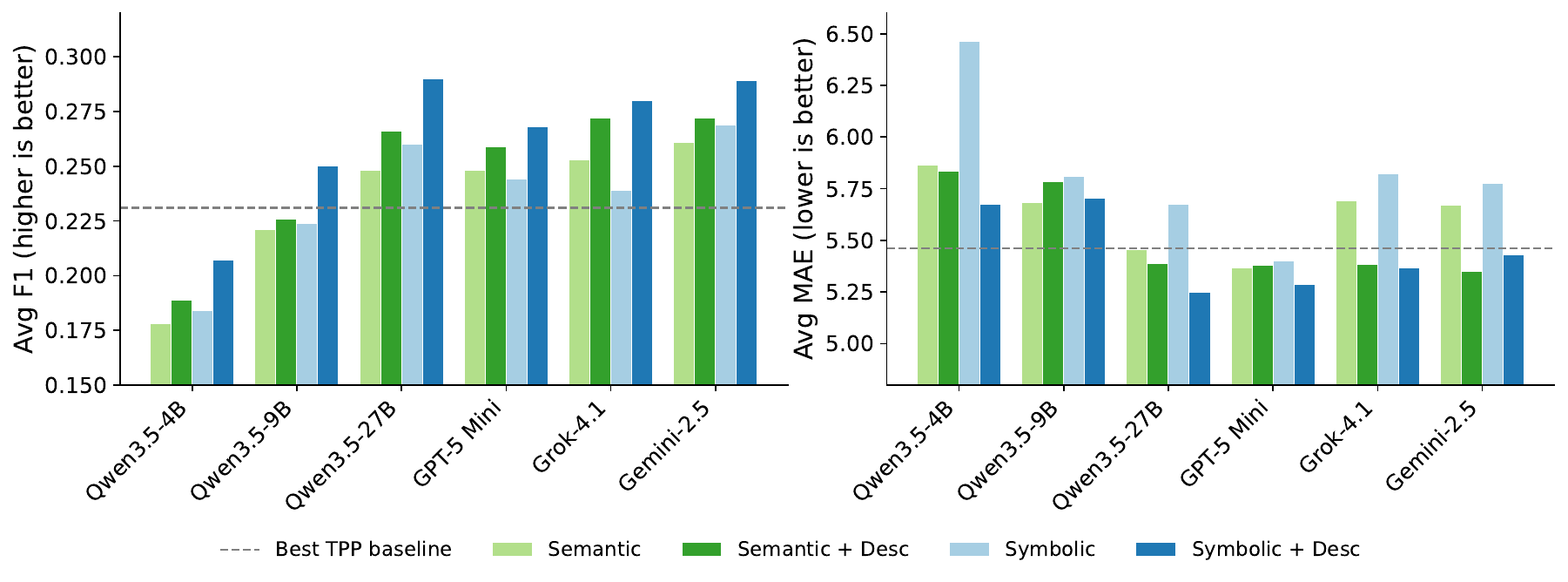}
  \caption{LLM Average performance across all datasets of \textsc{DescribeEvent}. Adding natural-language descriptions of events (as indicated by `+ Desc') consistently improves LLM performance, and pairing descriptions with using symbolic labels instead of semantic event names as prediction targets yields the largest gains. Dashed lines indicate the best non-LLM baseline for each metric.}
  \label{fig:summary}
\end{figure*}

We argue for the latter and identify two interface choices that, together, help LLMs use their pretrained knowledge for event prediction. First, we ground each event in a natural-language description. Standard event-prediction benchmarks represent events only as \((\texttt{event\_type}, \texttt{event\_time})\) pairs~\cite{xue2024easytpp}, stripping away the event semantics that LLMs could otherwise use. Yet in real settings, events often come with text that explains what happened, such as a commit message describing a code change. We therefore move to a more realistic setting: each event is supplemented with a text description that provides context and domain cues unavailable from categorical labels alone. This makes event prediction a natural inference task for LLMs, rather than a purely symbolic sequence-modeling problem.

The second choice is to decouple event semantics from the prediction space. Semantically loaded labels can bias LLMs toward pretrained label associations~\cite{zhao2021calibrate,reif2024quantifying} rather than the task context, an effect we call \emph{semantic bias}. We mitigate this by keeping event semantics in the input descriptions, while using symbolic labels for prediction. A symbolic label is a non-semantic identifier such as \texttt{type\_0} or \texttt{type\_1} that carries no label-name signal. This removes semantic bias and pushes an LLM to predict the next event from the descriptions of past events.
Together, these two choices expose event semantics where LLMs can use them and remove semantic bias where it hurts, effectively eliciting pretrained world knowledge for event prediction.

To evaluate whether these choices help LLMs predict the next event, we introduce \textsc{DescribeEvent}, a benchmark of $\sim$150K events drawn from seven datasets across six domains, with time scales from minutes to weeks. While prior datasets attach at most one description per sequence~\cite{xue2024easytpp,liu2024tppllm}, every event in \textsc{DescribeEvent} is paired with its own natural-language description. We evaluate six state-of-the-art LLMs under four settings, combining semantic or symbolic event labels with and without descriptions. We also compare against six non-LLM baselines.

Our experiments show that descriptions and symbolic labels are complementary: capable LLMs using both outperform description-free LLMs and non-LLM baselines on most datasets, as shown in Figure~\ref{fig:summary}. Descriptions with semantic labels capture only part of the gain, while symbolic labels remove semantic bias and unlock the full benefit. These gains are largest for models with stronger reasoning ability~\cite{rein2024gpqa,phan2025hle}, while weaker models gain little or sometimes degrade with richer context~\cite{cao2026context}.

In summary, our main contributions include:
\begin{itemize}
    \item We show that LLM knowledge is accessible for event prediction when the prediction interface exposes it properly.
    \item We identify per-event descriptions and symbolic labels as two interface choices that jointly produce the strongest and most consistent gains.
    \item We introduce \textsc{DescribeEvent}, a benchmark of $\sim$150K events across 7 datasets and 6 domains, and use it to show that the description benefit depends on model reasoning ability.
\end{itemize}

\section{Related Work}

We review prior work on temporal point processes, LLMs for time series, LLMs for event prediction, and the effect of label semantics on LLM predictions that motivates our use of symbolic labels.

\paragraph{Temporal Point Processes.}
Temporal point processes (TPPs) model sequences of discrete events in continuous time. The Hawkes process~\cite{hawkes1971spectra} introduced mutual excitation between event types and remains a widely used baseline for event prediction. Subsequent neural extensions replace parametric intensity functions with learned ones. RMTPP~\cite{du2016recurrent} uses recurrent networks to encode event history, and the Neural Hawkes Process~\cite{mei2017neural} models continuous-time dynamics with a neural architecture. More recent work applies self-attention to event sequences. For example, the Transformer Hawkes Process~\cite{zuo2020thp} and the Self-Attentive Hawkes Process~\cite{zhang2020sahp} replace recurrent encoders with attention mechanisms to model long-range dependencies. EasyTPP~\cite{xue2024easytpp} standardizes evaluation across public datasets and represents each event as a type--timestamp pair. This format is compact, but it omits the natural language descriptions that often accompany real-world events.

\paragraph{LLMs for Time Series.}
Large language models have recently been adapted to temporal data. Specifically, Time-LLM~\cite{jin2024timellm} reprograms LLMs for time series forecasting by aligning time series representations with language tokens. Chronos~\cite{ansari2024chronos} tokenizes time series values into a fixed vocabulary and trains transformer language models directly on them. Nevertheless, \citet{tan2024lmts} show that the LLM component can often be removed without hurting performance, challenging the assumption that pretrained language knowledge helps for temporal prediction. Similarly, \citet{Chang2024ACE} find that zero-shot LLMs lag behind conventional methods for temporal event forecasting. \citet{williams2025context} offer a counterpoint and introduce a benchmark where textual context is essential for time series forecasting. They show that their LLM prompting outperforms statistical and foundation model baselines.

\paragraph{LLMs for Event Prediction.}
In LLM-based event prediction, TPP-LLM~\cite{liu2024tppllm} fine-tunes LLMs on structured event records using the same type--timestamp format as classical TPPs. Several methods go further by incorporating textual information. LAMP~\cite{shi2023lamp} uses abductive reasoning with event metadata to let LLMs explain candidate future events proposed by a TPP predictor, Language-TPP~\cite{kong2026languagetpp} converts time intervals into byte-tokens that integrate descriptions with a language model, and LAST SToP~\cite{gupta2025laststop} designs specialized prompts with event descriptions for asynchronous time series. TESRBench~\cite{liu2025tesrbench} encodes event sequences and descriptions into a shared space for retrieval rather than next-event prediction. In a related direction, \citet{cao2026context} study whether additional context improves LLM temporal reasoning on a repurchase prediction task, and \citet{tan2025infer} study whether LLMs can infer event descriptions from time series rather than use descriptions to improve prediction. \citet{ye2024mirai} introduce MIRAI, a benchmark for evaluating LLM agents on international event forecasting, but focus on agent tool use within a single domain rather than on interface design across domains.

\paragraph{Label Semantics in LLMs.}
Prior work suggests that LLMs can carry non-trivial priors over label tokens. For example, \citet{zhao2021calibrate} show that in few-shot classification, LLMs tend to favor certain label strings regardless of the input, and propose calibration methods to correct this. \citet{reif2024quantifying} systematically measure and mitigate label bias across tasks. \citet{wei2023symbol} show that replacing natural language labels with arbitrary symbols, both during fine-tuning and at inference, improves in-context learning for NLP tasks. Our use of symbolic labels draws on this insight and extends it to event prediction, where symbolic event labels make descriptions the main source of event semantics rather than label priors.

\section{Approach}
\label{sec:approach}
In this section, we present two interface choices for improved LLM-based event prediction. We ground events in natural-language descriptions and decouple event semantics from prediction labels to reduce misleading pretrained label associations. Together, these choices expose useful context while preserving a cleaner prediction space.

\subsection{Preliminaries}
\label{sec:prelim}

\paragraph{Next-event prediction.}
An event sequence is an ordered list of events observed in time. In the standard representation used by prior event-prediction work~\cite{xue2024easytpp}, each event is a pair of a timestamp and a categorical type,
\begin{equation}
  e_i = (t_i, k_i),
\end{equation}
where $t_i \in \mathbb{R}$ is the timestamp and $k_i \in \mathcal{K}$ is a categorical type drawn from a finite vocabulary $\mathcal{K}$. A sequence of length $N$ is $S = (e_1, \ldots, e_N)$. The history prefix of length $n$ is $H_n = (e_1, \ldots, e_n)$, and the next-event prediction task is to produce $(t_{n+1}, k_{n+1})$ given $H_n$.

\paragraph{Prediction with an off-the-shelf LLM.}
We serialize each event into text as ``$[k_i]\ {+}\Delta t_i$'', where $\Delta t_i = t_i - t_{i-1}$ is the inter-event gap~\cite{liu2024tppllm,kong2026languagetpp}. The prompt pairs this serialized history with a system message, which states the task, lists the valid labels in $\mathcal{K}$, and asks the model to reason before returning the next event as \{\texttt{event\_type}, \texttt{time}\}. At step $n$, the model produces
\begin{equation}
  (r_{n+1}, \hat{t}_{n+1}, \hat{k}_{n+1}) = \mathrm{LLM}(H_n),
\end{equation}
where $r_{n+1}$ is model reasoning. We follow teacher forcing by appending the ground-truth $e_{n+1}$ to the history before the next prediction~\cite{mei2017neural,xue2024easytpp,gupta2025laststop}.

\subsection{Text Descriptions}
\label{sec:approach-desc}

Event types alone (typically used for event prediction) do not carry enough semantic content for LLMs to leverage their pretrained knowledge. An event type like \texttt{push} is merely a name with no domain context behind it. Without that context, the model has little information to ground its pretrained knowledge on, and works only from the label tokens themselves.

A per-event description surfaces this context in the model's input. We extend each event with a natural-language description $d_i$, so an event becomes a triple
\begin{equation}
  e_i = (t_i, k_i, d_i).
\end{equation}
For a code-repository event, $d_i$ can mention the commit message, the number of commits, or the files changed. For a seismic event, it can mention the magnitude, depth, or location.

Descriptions enter only as input context, and the prediction target stays $(t_{n+1}, k_{n+1})$ as in Section~\ref{sec:prelim}. With this added context, each event in the history reads as text rather than the label name alone, so the model can apply its pretrained world knowledge to next-event prediction.

\subsection{Removing Semantic Bias}
\label{sec:approach-symbolic}

\begin{figure}[t]
  \centering
  \includegraphics[width=\columnwidth]{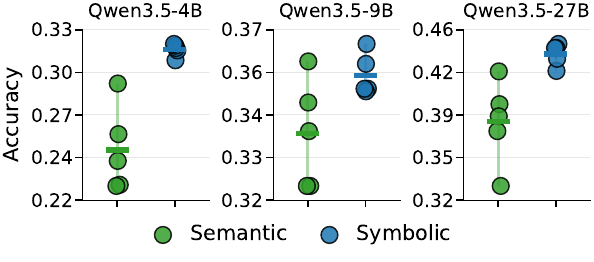}
  \caption{Accuracy under random label permutations. Semantic labels are permutation-sensitive, while symbolic labels are not.}
  \label{fig:cross-size-accuracy}
\end{figure}

Event type names are themselves natural-language tokens, and an LLM can carry sequential associations between such names from pretraining text~\cite{zhao2021calibrate,reif2024quantifying}. For example, in code-repository text \texttt{push} often appears near \texttt{merge}, so an LLM expects \texttt{merge} after \texttt{push} before reading any event history. We call this effect \emph{semantic bias}: the model predicts the next event from these pretrained label-name associations, not from the actual past events. As a result, the model can produce a plausible-sounding prediction from labels alone, ignoring the description even when the description carries decisive information.

We run a label-permutation test to expose semantic bias. Specifically, we omit descriptions so that the label sequence is the only source of signal, and we randomly relabel the type vocabulary across runs while leaving the underlying event sequence unchanged. For example, under semantic labels, \texttt{push} might be relabeled as \texttt{pull} and \texttt{commit} as \texttt{push}. Under symbolic labels, \texttt{push} maps to \texttt{type\_1} in one run and \texttt{type\_4} in another. As shown in Figure~\ref{fig:cross-size-accuracy}, symbolic accuracy is more stable than semantic accuracy across permutations. This gap reveals semantic bias, since the label names themselves carry predictive signal beyond structure.

To prevent semantic bias from competing with descriptions, we adopt a fixed bijective mapping $\ell_{\mathrm{sym}}$ that rewrites each semantic type as an anonymous token. Specifically, the $i$-th event type in the vocabulary $\mathcal{K}$ is renamed to the string \texttt{type\_i}, so \texttt{push} and \texttt{commit} become \texttt{type\_0} and \texttt{type\_1}. With semantic bias removed, the model is encouraged to rely on the description $d_i$ to predict the next event.

\section{The \textsc{DescribeEvent} Benchmark}
\label{sec:benchmark}

\begin{figure}[t]
  \centering
  \includegraphics[width=0.98\columnwidth]{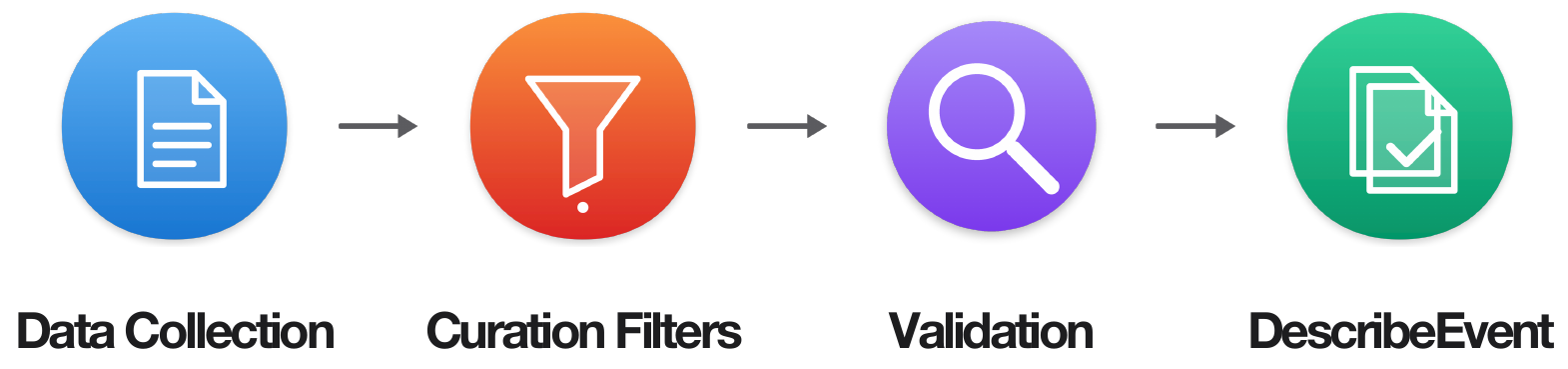}
  \caption{\textsc{DescribeEvent} data curation pipeline.}
  \label{fig:data-pipeline}
\end{figure}

We introduce \textsc{DescribeEvent}, a benchmark that pairs every event with a natural-language description, to test whether LLMs can use these descriptions for next-event prediction. We first describe the data collection, then the curation filters, and finally the validation diagnostics and dataset statistics. Figure~\ref{fig:data-pipeline} summarizes the full pipeline.

\subsection{Data Collection}

\begin{figure}[t]
  \centering
  \includegraphics[width=0.9\columnwidth]{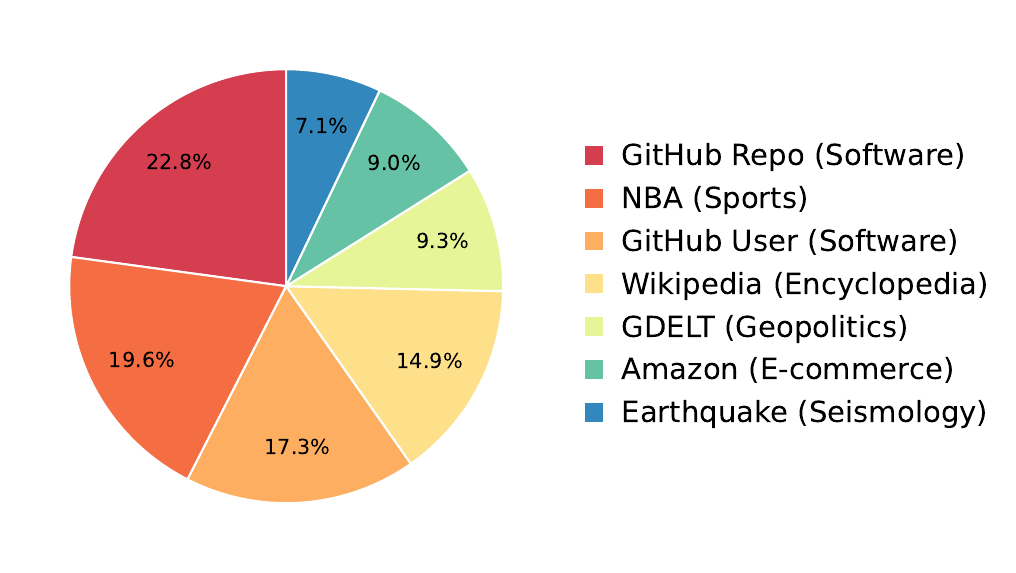}
  \caption{Dataset distribution of \textsc{DescribeEvent} by total events. The benchmark spans seven datasets across six diverse domains.}
  \label{fig:domain-dist}
\end{figure}

\begin{figure}[t]
  \centering
  \includegraphics[width=0.98\columnwidth]{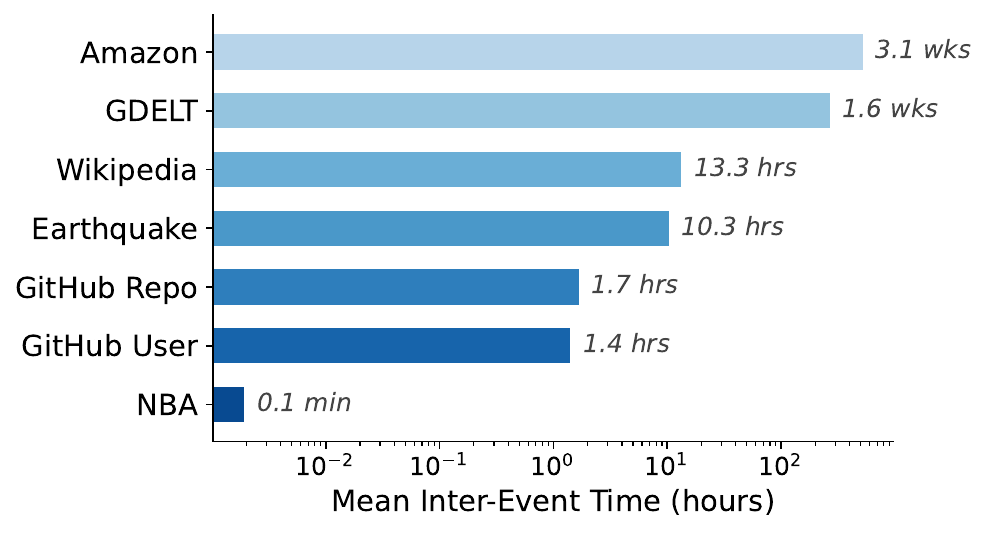}
  \caption{Mean inter-event time across datasets. \textsc{DescribeEvent} covers temporal granularities from minutes to weeks.}
  \label{fig:temporal-scale}
\end{figure}

\textsc{DescribeEvent} covers six domains across seven datasets. As shown in Figure~\ref{fig:domain-dist}, it includes Amazon product reviews, USGS earthquake records, GDELT geopolitical interactions, NBA play-by-play, Wikipedia article edits, and GitHub activity under two different views, per repository and per user. These domains span a wide range of actors, from individuals to institutions to natural processes. The temporal granularities range from minutes to weeks, as shown in Figure~\ref{fig:temporal-scale}. Details about the dataset sources and prediction tasks are provided in Table~\ref{tab:dataset-overview} in Appendix~\ref{app:dataset-stats}.

Every event extends the EasyTPP format~\cite{xue2024easytpp} with one text field for the description. Specifically, we inherit the timestamp $t_i$ and event type $k_i$ for each event, and add the natural-language description $d_i$ defined in Section~\ref{sec:approach-desc}.
Each description $d_i$ is taken from the raw source text attached to the event at the time it occurred, mirroring a real-time collection setting: when an event and its description are recorded, later events have not yet happened. This temporal ordering prevents information about future events from entering earlier descriptions.
The label space for $k_i$ differs across datasets and is listed in Appendix~\ref{app:label-space}.

\subsection{Curation Filters}
\label{sec:curation-filters}

We apply three universal filters to every dataset to improve the quality of each sequence, since a high-quality sequence should not be solvable by trivial baselines, such as always predicting the most common type or exploiting long runs of identical events. The three filters we applied are a \emph{dominant-type cap} on the fraction of events sharing a single type, a \emph{consecutive-repeat cap} on the fraction of adjacent events sharing the same type, and a \emph{minimum type diversity} on the number of distinct types per sequence.

Additionally, each dataset passes through domain-specific filters to further improve the quality of the sequences, such as magnitude thresholds for earthquakes and bot-account exclusion for GitHub. Those per-dataset thresholds are detailed in Appendix~\ref{app:filter-thresholds}.

\subsection{Validation and Statistics}
\label{sec:benchmark-stats}

We validate each dataset with two diagnostics. The first is \emph{Recoverability}, denoted by $R$. It is the accuracy of a validator LLM~\cite{gilardi2023chatgpt} predicting the type from the description alone. High $R$ shows that the description is informative enough to identify the event type. As shown in Table~\ref{tab:description-validation}, $R$ exceeds random guess by a large margin on every dataset.

To check that a high $R$ does not come from descriptions that merely paraphrase the labels, we report a second diagnostic, \emph{Similarity}. For each description, we compute two cosine similarities under a sentence encoder~\cite{reimers-gurevych-2019-sentence}, $\cos_{\text{true}}$ against its true label name and $\cos_{\text{other}}$ against the other labels. Descriptions that paraphrase their labels would yield a large gap. As shown in Table~\ref{tab:description-validation}, this gap is small on every dataset. Together with the large recoverability gain above, this shows that descriptions in \textsc{DescribeEvent} carry event content beyond what the label name already provides, as intended in Section~\ref{sec:approach-desc}.

After the above curation processes, \textsc{DescribeEvent} contains 148{,}185 events organized into 1{,}940 sequences across seven datasets. Detailed dataset statistics are reported in Table~\ref{tab:dataset-stats}.

\begin{table}[!t]
  \centering
  \resizebox{\columnwidth}{!}{%
  \begin{tabular}{lcccccc}
  \toprule
   & \multicolumn{3}{c}{\textbf{Recoverability}} & \multicolumn{3}{c}{\textbf{Similarity}} \\
  \cmidrule(lr){2-4} \cmidrule(lr){5-7}
  \textbf{Dataset} & Random & $R$ & $\Delta_{\text{gain}}$ & $\cos_{\text{other}}$ & $\cos_{\text{true}}$ & $\Delta_{\text{leak}}$ \\
  \midrule
  GitHub Repo  & 0.125 & 0.905 & 0.780 & 0.114 & 0.187 & 0.073 \\
  GitHub User  & 0.125 & 0.897 & 0.772 & 0.122 & 0.228 & 0.106 \\
  Amazon       & 0.100 & 0.904 & 0.804 & 0.155 & 0.358 & 0.203 \\
  Earthquake   & 0.125 & 0.884 & 0.759 & 0.257 & 0.388 & 0.131 \\
  GDELT        & 0.143 & 0.611 & 0.468 & 0.112 & 0.227 & 0.115 \\
  NBA          & 0.111 & 0.992 & 0.881 & 0.249 & 0.399 & 0.150 \\
  Wikipedia    & 0.167 & 0.602 & 0.435 & 0.233 & 0.351 & 0.118 \\
  \bottomrule
  \end{tabular}}
  \caption{Description validation diagnostics across the seven datasets in \textsc{DescribeEvent}. $\Delta_{\text{gain}}$ measures recovery above chance, where higher is better. $\Delta_{\text{leak}}$ measures label leakage, where lower is better and indicates that descriptions do not paraphrase labels.}
  \label{tab:description-validation}
\end{table}

\begin{table*}[!t]
  \centering
  \resizebox{\textwidth}{!}{%
  \begin{tabular}{lcccccccccccccc}
  \toprule
  & \multicolumn{2}{c}{\textbf{GitHub Repo}} & \multicolumn{2}{c}{\textbf{GitHub User}} & \multicolumn{2}{c}{\textbf{Amazon}} & \multicolumn{2}{c}{\textbf{Earthquake}} & \multicolumn{2}{c}{\textbf{GDELT}} & \multicolumn{2}{c}{\textbf{NBA}} &
  \multicolumn{2}{c}{\textbf{Wikipedia}} \\
  \cmidrule(lr){2-3} \cmidrule(lr){4-5} \cmidrule(lr){6-7} \cmidrule(lr){8-9} \cmidrule(lr){10-11} \cmidrule(lr){12-13} \cmidrule(lr){14-15}
  \textbf{Model} & F1$\uparrow$ & MAE$\downarrow$ & F1$\uparrow$ & MAE$\downarrow$ & F1$\uparrow$ & MAE$\downarrow$ & F1$\uparrow$ & MAE$\downarrow$ & F1$\uparrow$ & MAE$\downarrow$ & F1$\uparrow$ & MAE$\downarrow$ & F1$\uparrow$ & MAE$\downarrow$ \\
  \midrule
  \rowcolor{gray!15} \multicolumn{15}{c}{\textit{Semantic Event Type}} \\
  \emph{Qwen3.5-4B} & 0.226 & 3.002 & 0.206 & 2.327 & 0.140 & 6.728 & 0.150 & 0.406 & 0.141 & 10.303 & 0.191 & 0.128 & 0.190 & 18.171 \\
  \emph{Qwen3.5-9B} & 0.290 & 2.832 & 0.245 & 1.830 & 0.148 & 6.622 & 0.167 & 0.370 & 0.133 & 10.409 & 0.262 & 0.092 & 0.304 & 17.634 \\
  \emph{Qwen3.5-27B} & 0.349 & 2.334 & 0.305 & 1.563 & 0.159 & 5.959 & 0.225 & 0.365 & 0.137 & 10.188 & 0.278 & 0.082 & 0.280 & 17.696 \\
  \emph{GPT-5 Mini} & 0.332 & 2.252 & 0.317 & 1.473 & 0.131 & 6.012 & 0.220 & 0.344 & 0.146 & 10.139 & 0.281 & 0.075 & 0.311 & 17.268 \\
  \emph{Grok-4.1} & 0.342 & 2.487 & 0.328 & 1.626 & 0.149 & 6.612 & 0.187 & 0.345 & 0.168 & 10.112 & 0.307 & 0.077 & 0.287 & 18.585 \\
  \emph{Gemini-2.5} & 0.351 & 2.295 & 0.309 & 1.585 & 0.168 & 5.957 & 0.231 & 0.350 & 0.156 & 10.201 & 0.320 & 0.077 & 0.292 & 19.233 \\
  \midrule
  \rowcolor{gray!15} \multicolumn{15}{c}{\textit{Symbolic Event Type}} \\
  \emph{Qwen3.5-4B} & 0.216 & 3.709 & 0.232 & 2.893 & 0.117 & 7.005 & 0.174 & 0.379 & 0.150 & 10.715 & 0.163 & 0.482 & 0.233 & 20.061 \\
  \emph{Qwen3.5-9B} & 0.281 & 2.947 & 0.265 & 1.990 & 0.157 & 6.507 & 0.214 & 0.402 & 0.161 & 10.748 & 0.216 & 0.438 & 0.271 & 17.647 \\
  \emph{Qwen3.5-27B} & 0.352 & 2.422 & 0.327 & 1.570 & 0.161 & 5.975 & 0.235 & 0.367 & 0.149 & 10.816 & 0.282 & 0.250 & 0.315 & 18.323 \\
  \emph{GPT-5 Mini} & 0.336 & 2.353 & 0.301 & 1.537 & 0.142 & 6.021 & 0.211 & 0.348 & 0.148 & 10.187 & 0.264 & 0.094 & 0.305 & 17.268 \\
  \emph{Grok-4.1} & 0.358 & 2.518 & 0.306 & 1.669 & 0.139 & 6.815 & 0.175 & 0.353 & 0.150 & 10.556 & 0.285 & 0.086 & 0.261 & 18.749 \\
  \emph{Gemini-2.5} & 0.349 & 2.329 & 0.338 & 1.568 & 0.183 & 6.123 & 0.229 & 0.356 & 0.163 & 10.597 & 0.308 & 0.083 & 0.311 & 19.372 \\
  \midrule
  \rowcolor{gray!15} \multicolumn{15}{c}{\textit{Semantic Event Type \textbf{w/ Description}}} \\
  \emph{Qwen3.5-4B} & 0.229 & 2.844 & 0.182 & 1.949 & 0.127 & 6.095 & 0.207 & 0.397 & 0.120 & 12.057 & 0.236 & 0.106 & 0.225 & 17.406 \\
  \emph{Qwen3.5-9B} & 0.307 & 2.724 & 0.246 & 1.790 & 0.111 & 6.600 & 0.194 & 0.415 & 0.156 & 10.706 & 0.282 & 0.089 & 0.289 & 18.181 \\
  \emph{Qwen3.5-27B} & 0.353 & 2.332 & 0.288 & 1.489 & 0.160 & 5.806 & 0.219 & 0.359 & 0.160 & 10.392 & 0.317 & 0.078 & 0.362 & 17.268 \\
  \emph{GPT-5 Mini} & 0.330 & 2.445 & 0.277 & 1.647 & 0.157 & 5.841 & 0.252 & 0.345 & 0.151 & 10.087 & 0.316 & 0.072 & 0.330 & 17.206 \\
  \emph{Grok-4.1} & 0.344 & 2.459 & 0.285 & 1.576 & 0.169 & 6.084 & 0.238 & \textbf{0.334} & 0.158 & 10.073 & 0.329 & 0.072 & 0.378 & 17.082 \\
  \emph{Gemini-2.5} & 0.333 & 2.332 & 0.283 & 1.489 & \textbf{0.187} & 5.827 & 0.240 & 0.336 & 0.173 & \textbf{9.539} & 0.359 & 0.072 & 0.329 & 17.852 \\
  \midrule
  \rowcolor{gray!15} \multicolumn{15}{c}{\textit{Symbolic Event Type \textbf{w/ Description}}} \\
  \emph{Qwen3.5-4B} & 0.263 & 2.452 & 0.222 & 1.808 & 0.155 & 5.922 & 0.185 & 0.406 & 0.168 & 10.985 & 0.227 & 0.112 & 0.229 & 18.024 \\
  \emph{Qwen3.5-9B} & 0.310 & 2.774 & 0.261 & 1.700 & 0.164 & 6.118 & 0.206 & 0.391 & 0.173 & 10.266 & 0.294 & 0.080 & 0.340 & 18.601 \\
  \emph{Qwen3.5-27B} & \textbf{0.383} & 2.277 & 0.340 & \textbf{1.453} & 0.171 & 5.801 & \textbf{0.255} & 0.362 & 0.158 & 10.154 & 0.344 & 0.074 & \textbf{0.379} & \textbf{16.615} \\
  \emph{GPT-5 Mini} & 0.372 & \textbf{2.237} & \textbf{0.342} & 1.459 & 0.149 & \textbf{5.758} & 0.235 & 0.348 & 0.156 & 10.058 & 0.288 & 0.077 & 0.331 & 17.056 \\
  \emph{Grok-4.1} & 0.355 & 2.365 & 0.323 & 1.526 & 0.177 & 6.087 & 0.242 & \textbf{0.334} & 0.156 & 10.289 & 0.341 & \textbf{0.071} & 0.365 & 16.893 \\
  \emph{Gemini-2.5} & 0.361 & 2.295 & 0.324 & 1.492 & 0.174 & 5.851 & 0.242 & 0.349 & \textbf{0.182} & 10.080 & \textbf{0.369} & \textbf{0.071} & 0.368 & 17.864 \\
  \bottomrule
  \end{tabular}}
  \caption{LLM performance on \textsc{DescribeEvent}. Best results per dataset are in \textbf{bold}.}
  \label{tab:main results}
\end{table*}

\section{Experiments}
\label{sec:experiments}

\subsection{Experimental Setup}

\paragraph{Choice of LLMs.}
In our main experiments, we evaluate six large language models that span open and closed weights and a range of parameter scales. From the open-source Qwen3.5 family~\cite{qwen3report}, we include three sizes at 4B, 9B, and 27B parameters. We also include three close-source models, GPT-5 Mini~\cite{openai2025gpt5}, Grok-4.1 Fast~\cite{xai2025grok41}, and Gemini-2.5 Flash~\cite{comanici2025gemini25}. In our analysis, we probe how reasoning ability shapes the benefit of descriptions, and additionally include two less reasoning-capable models, Llama-3-8B~\cite{grattafiori2024llama3} and GPT-4o-mini~\cite{openai2024gpt4omini}.

\paragraph{Inference with LLMs.}
We evaluate the four settings defined by two binary factors (with/without description and symbolic/semantic labels) as described in Section~\ref{sec:approach}. We extend the base serialization from Section~\ref{sec:prelim} so that each event is written as ``$[k_i]\ {+}\Delta t_i\ |\ d_i$''. In the symbolic setting, $k_i$ is replaced by $\ell_{\mathrm{sym}}(k_i)$, and in the description-off settings, ``$|\ d_i$'' is removed. All four settings share the same prompting protocol, including the brief reasoning step before the final answer, so they differ only in how events are represented. One example per setting is given in Figure~\ref{fig:prompts} in Appendix~\ref{app:prompts}. Additional inference configuration is detailed in Appendix~\ref{app:inference}.

\paragraph{Evaluation Metrics.}
We evaluate predictions on next event type and time. For type, we report the macro-averaged F1 over the type vocabulary $\mathcal{K}$,
\begin{equation}
  \text{Macro-F1} = \frac{1}{|\mathcal{K}|} \sum_{k \in \mathcal{K}} \mathrm{F1}_k,
\end{equation}
where $\mathrm{F1}_k$ is the harmonic mean of precision and recall for type $k$. For time, we report Mean Absolute Error on the predicted inter-event gap, in each dataset's native time unit as detailed in Table~\ref{tab:dataset-stats},
\begin{equation}
  \mathrm{MAE} = \frac{1}{|\mathcal{P}|} \sum_{(s,n) \in \mathcal{P}} \bigl|\widehat{\Delta t}^{(s)}_{n+1} - \Delta t^{(s)}_{n+1}\bigr|,
\end{equation}
where $\mathcal{P}$ is the set of (sequence, position) pairs predicted across the test set.

\begin{figure*}[!t]
  \centering
  \includegraphics[width=\textwidth]{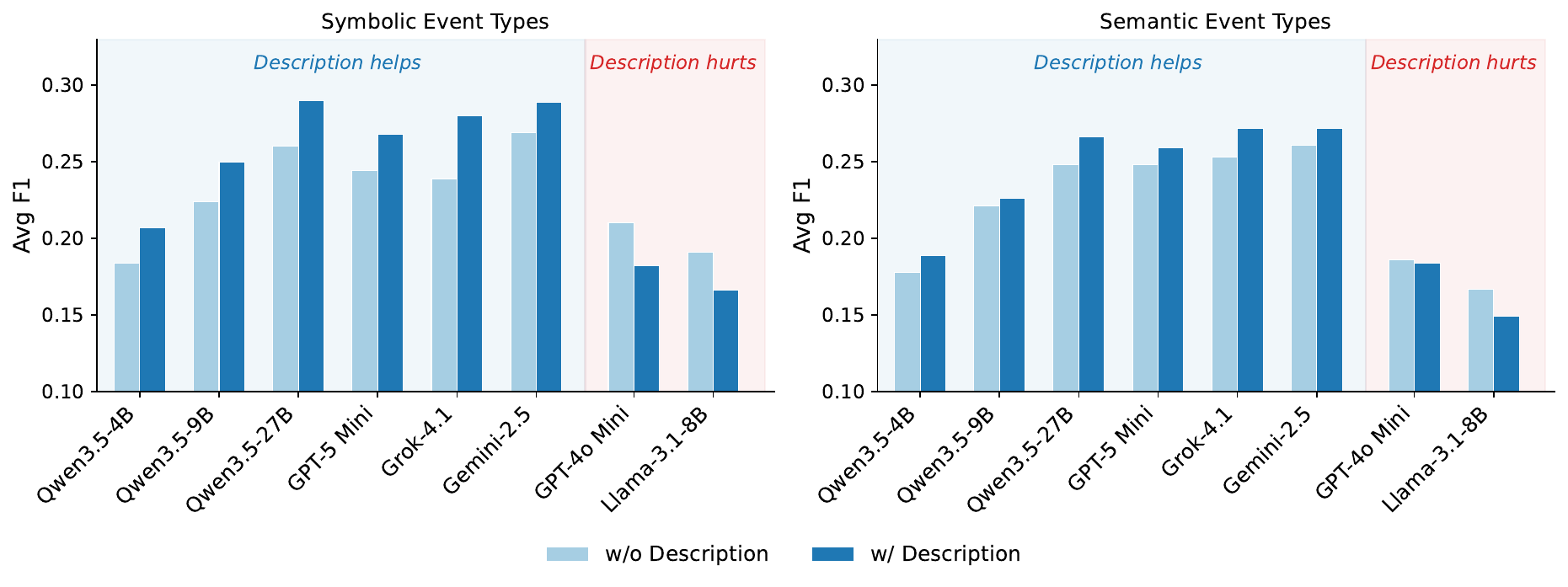}
  \caption{Effect of adding event descriptions on Avg F1 under Symbolic and Semantic label settings. Models with strong reasoning capability benefit from descriptions, while less capable models see degraded performance.}
  \label{fig:desc-effect}
\end{figure*}

\subsection{Performance of Large Language Models}

Table~\ref{tab:main results} reports per-dataset F1 and MAE for each LLM on \textsc{DescribeEvent}, and Figure~\ref{fig:summary} shows each model's average across the seven datasets.

We first observe that adding descriptions to semantic event labels generally improves performance. Table~\ref{tab:main results} shows that F1 increases and MAE drops for nearly every LLM when descriptions are added to semantic labels. This indicates that descriptions give the model information beyond what types and timestamps alone provide, even when semantic bias is still present.
Removing label semantics by using symbolic event types helps suppress this bias, but without descriptions it also removes the semantic cues that the LLM could use. The model is then left mainly with sequence patterns in labels and timestamps, which explains why symbolic labels alone perform similarly to semantic labels for most models.

We also observe that descriptions are most effective when paired with symbolic labels, achieving the best per-dataset F1 and MAE on most datasets in Table~\ref{tab:main results} and the highest average F1 and lowest average MAE in Figure~\ref{fig:summary}. These results support the semantic bias intuition from Section~\ref{sec:approach-symbolic}: when label names no longer provide semantic cues, the model is encouraged to use the event descriptions, allowing the information they carry, and any pretrained knowledge they activate, to contribute more directly to prediction.

\begin{figure}[t]
  \centering
  \includegraphics[width=0.8\columnwidth]{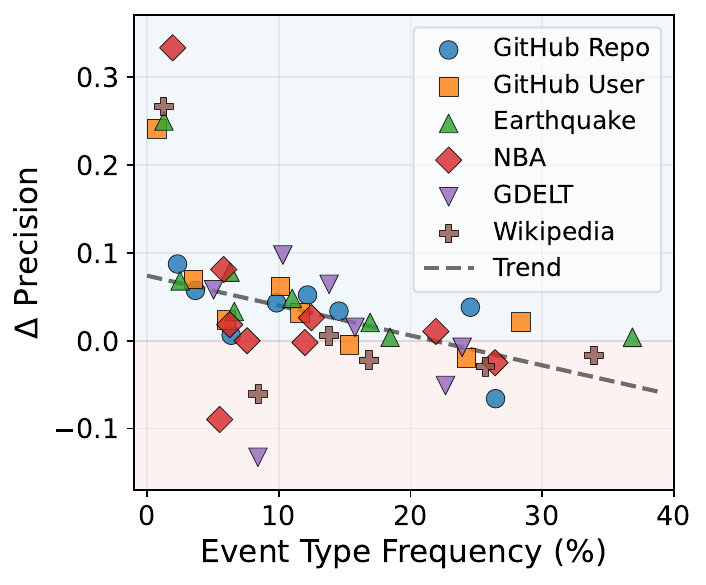}
  \caption{Precision gain from using Symbolic over Semantic labels plotted against event type frequency.}
  \label{fig:rare-event}
\end{figure}

\subsection{Description Benefit Tracks Reasoning Capability}
\label{sec:reasoning}

Predicting from a description is itself a non-trivial reasoning task, since the model must extract causal signal from free-form text and integrate it with the temporal pattern of past events. We thus examine the relationship between a model's reasoning capability and its ability to use descriptions. We use two reasoning benchmarks as proxies\footnote{Benchmark scores are sourced from Artificial Analysis, \url{https://artificialanalysis.ai/}.} : GPQA Diamond~\cite{rein2024gpqa} tests graduate-level domain reasoning, and Humanity's Last Exam (HLE)~\cite{phan2025hle} tests multi-disciplinary multi-step reasoning.

Figure~\ref{fig:desc-effect} compares average F1 with and without descriptions and reveals a split. And as shown in Table~\ref{tab:gpqa} in Appendix~\ref{app:gpqa}, Models that benefit from descriptions score at least 77\% on GPQA Diamond and at least 7\% on HLE, while GPT-4o Mini and Llama-3.1-8B perform significantly worse on both and see overall F1 drop when descriptions are added. We hypothesize that without strong reasoning capabilities, these weaker models are distracted by the description and lose the temporal patterns they would otherwise follow~\cite{wu2024distraction}. Appendix~\ref{app:small-models} and Table~\ref{tab:small-models} provide further details about the performance of the weaker models.

\begin{table*}[!t]
  \centering
  \resizebox{0.96\textwidth}{!}{%
  \begin{tabular}{lcccccccccccccc}
  \toprule
  & \multicolumn{2}{c}{\textbf{GitHub Repo}} & \multicolumn{2}{c}{\textbf{GitHub User}} & \multicolumn{2}{c}{\textbf{Amazon}} & \multicolumn{2}{c}{\textbf{Earthquake}} & \multicolumn{2}{c}{\textbf{GDELT}} & \multicolumn{2}{c}{\textbf{NBA}} &
  \multicolumn{2}{c}{\textbf{Wikipedia}} \\
  \cmidrule(lr){2-3} \cmidrule(lr){4-5} \cmidrule(lr){6-7} \cmidrule(lr){8-9} \cmidrule(lr){10-11} \cmidrule(lr){12-13} \cmidrule(lr){14-15}
  \textbf{Model} & F1$\uparrow$ & MAE$\downarrow$ & F1$\uparrow$ & MAE$\downarrow$ & F1$\uparrow$ & MAE$\downarrow$ & F1$\uparrow$ & MAE$\downarrow$ & F1$\uparrow$ & MAE$\downarrow$ & F1$\uparrow$ & MAE$\downarrow$ & F1$\uparrow$ & MAE$\downarrow$ \\
  \midrule
  \rowcolor{gray!15} \multicolumn{15}{c}{\textit{non-LLM Baselines}} \\
  \emph{ARIMA} & 0.166 & 3.075 & 0.193 & 2.152 & \textbf{0.130} & 7.010 & 0.149 & 0.374 & 0.151 & 10.363 & 0.059 & 0.105 & 0.175 & 24.561 \\
  \emph{KNN} & \textbf{0.332} & 2.906 & \textbf{0.327} & 2.244 & 0.117 & 6.875 & \textbf{0.175} & 0.397 & \textbf{0.153} & 11.139 & 0.260 & 0.089 & \textbf{0.256} & 21.239 \\
  \emph{Hawkes} & 0.067 & 4.215 & 0.109 & 3.805 & 0.056 & 6.427 & 0.067 & 0.561 & 0.055 & 18.596 & 0.046 & 0.154 & 0.084 & 102.286 \\
  \emph{NHP} & 0.159 & 2.462 & 0.131 & 1.665 & 0.044 & 5.934 & 0.148 & 0.343 & 0.095 & 10.772 & 0.148 & 0.105 & 0.084 & 18.227 \\
  \emph{RMTPP} & 0.258 & 2.688 & 0.298 & 1.944 & 0.026 & 6.326 & 0.161 & 0.360 & 0.058 & \textbf{9.688} & 0.338 & \textbf{0.073} & 0.138 & 23.022 \\
  \emph{MLP} & 0.307 & 2.637 & 0.287 & 1.903 & 0.048 & 6.501 & 0.163 & 0.359 & 0.066 & 10.306 & 0.329 & 0.080 & 0.090 & 24.188 \\
  \midrule
  \rowcolor{gray!15} \multicolumn{15}{c}{\textit{non-LLM Baselines \textbf{w/ Description}}} \\
  \emph{NHP} & 0.131 & \textbf{2.288} & 0.117 & \textbf{1.563} & 0.034 & \textbf{5.780} & 0.151 & \textbf{0.335} & 0.138 & 11.423 & 0.078 & 0.115 & 0.177 & \textbf{16.720} \\
  \emph{RMTPP} & 0.248 & 2.944 & 0.227 & 2.075 & 0.028 & 6.339 & 0.161 & 0.351 & 0.109 & 10.265 & \textbf{0.362} & 0.079 & 0.199 & 20.202 \\
  \emph{MLP} & 0.196 & 2.733 & 0.180 & 2.149 & 0.043 & 5.982 & 0.145 & 0.359 & 0.115 & 10.406 & 0.347 & 0.075 & 0.097 & 23.427 \\
  \bottomrule
  \end{tabular}}
  \caption{Per-dataset performance of non-LLM baselines. Best result per dataset is in \textbf{bold}.}
  \label{tab:tpp-baselines}
\end{table*}

\subsection{Symbolic Labels Improve Rare Event Precision}
\label{sec:rare-event}

We next investigate why symbolic labels outperform semantic labels when both are paired with descriptions. Our intuition follows the semantic bias defined in Section~\ref{sec:approach-symbolic}: an LLM may predict an event type from its label name alone even when the description does not support that prediction~\cite{mccoy2024embers}. This can create bias-driven false positives. Symbolic labels remove the semantic signal from the label name, encouraging the model to base its prediction on the description instead. Since precision is $\text{TP}_t/(\text{TP}_t+\text{FP}_t)$, the same number of false positives has a smaller effect for common events, where there are many true positives, but a much larger effect for rare events, where the true-positive base is small. Therefore, if symbolic labels reduce semantic-bias-driven false positives, the precision gain should be strongest for rare events.

To test this hypothesis, we compute performance differences between symbolic and semantic labels, with descriptions included in both settings. For each event type, we measure the change in precision, recall, and F1, and compare it with that type's frequency within its dataset. Figure~\ref{fig:rare-event} shows the precision results, while Figure~\ref{fig:rare-event-all} in Appendix~\ref{app:rare-event-all} shows recall and F1. We exclude Amazon because its large label space would dominate the results. Across the remaining 46 event types, the precision gain from symbolic labels increases sharply as frequency decreases, with a Pearson correlation of $r=-0.51$ ($p\approx 0.0003$). In contrast, recall shows almost no relationship with frequency ($r=-0.01$), suggesting that symbolic labels mainly reduce false positives rather than changing how often true events are recovered. This does not mean that rare events carry stronger semantic bias. Prior work links label bias to properties of the label tokens and their pretraining distribution, rather than to a label's frequency in a downstream dataset~\cite{zhao2021calibrate,reif2024quantifying}. Thus, the frequency effect is better explained by precision: rare events have fewer true positives, so the same number of false positives has a larger impact.

\subsection{Comparison with non-LLM Baselines}
\label{sec:tpp-baselines}

We compare our approach with six baselines spanning statistical, nonparametric, temporal point process (TPP), and neural sequence-modeling approaches. The TPP baselines include one classical model, Hawkes Process~\cite{hawkes1971spectra}, and two neural TPPs: NHP~\cite{mei2017neural} and RMTPP~\cite{du2016recurrent}. We also include ARIMA~\cite{box2015time} and KNN as non-TPP baselines, and an MLP as a simple neural sliding-window baseline. We use no descriptions for ARIMA, KNN, and Hawkes. For the neural baselines, NHP, RMTPP, and MLP, we additionally evaluate description-aware variants that encode each event's description with Sentence-BERT~\cite{reimers-gurevych-2019-sentence} and concatenate the embedding with the type and time features at the model input. We exclude fine-tuned LLM event predictors~\cite{liu2024tppllm,kong2026languagetpp,gupta2025laststop} because they require a large training corpus, which lies beyond the scope of our evaluation.

ARIMA forecasts inter-event times with small-order models chosen by grid search and predicts the most frequent type as the label. KNN matches the most similar sliding windows in the sequence's history by type and temporal similarity. Hawkes Process models conditional intensity as background rates with exponentially decaying excitation from past events, and predicts the highest-intensity type along with the reciprocal of total intensity as the time. NHP is a neural TPP based on a continuous-time LSTM whose hidden state decays between events, while RMTPP is a neural TPP based on a GRU with separate output heads for type and time. Finally, MLP is a two-layer feedforward network over a fixed sliding window of recent events. We train the parametric baselines on the first 80\% to 90\% events in each sequence (same length that our LLM method receives in its prompt), and evaluate autoregressively on the remaining events.

Table~\ref{tab:tpp-baselines} provides the non-LLM comparison point for our results. Overall, the strongest LLM setting in Table~\ref{tab:main results}, which combines descriptions with symbolic labels, outperforms the best non-LLM baselines on most datasets for both metrics. Among the non-LLM methods, KNN is the strongest baseline for event-type prediction, while description-aware NHP is strongest for time prediction. The effect of descriptions on non-LLM baselines is mixed: adding Sentence-BERT embeddings often improves MAE, but does not consistently improve F1. This suggests that description embeddings help these baselines capture timing information more reliably than event-type semantics.

\section{Conclusion}

In this paper, we showed that LLM world knowledge can improve event prediction when the prediction interface exposes it. We identified two key interface choices: grounding events in natural-language descriptions and using symbolic labels as prediction targets. Their combination yields the strongest results, with capable LLMs outperforming description-free variants and non-LLM baselines on most datasets, while weaker models benefit little or even degrade. To support this evaluation, we introduce \textsc{DescribeEvent}, a benchmark of 150K events across seven datasets and six domains, and show that the benefit of descriptions is closely tied to model reasoning ability.

\section{Limitations}
Our work has several limitations. First, \textsc{DescribeEvent} is designed as an evaluation benchmark; we do not explore using it as training data to fine-tune LLMs for description-grounded prediction, e.g., via parameter-efficient adaptation~\cite{hu2022lora, wu2026ulpt,wu2026tokmem}, which is a promising direction for future work. Second, while we identify a correlation between reasoning capability, as measured by GPQA Diamond, and description utilization, establishing a causal mechanism requires further investigation. Third, LLM inference may take seconds per prediction versus tens of milliseconds for specialized event-prediction non-LLM models in our experiments; we view this added latency as a trade-off for better performance and the interpretability offered by the model's natural-language reasoning. Finally, we represent inter-arrival times as raw floating-point numbers in text using standard tokenizers while improving their representations is worth exploring further~\cite{singh2024tokenizationcountsimpacttokenization}.

\bibliography{custom}

\clearpage
\appendix

\section{Datasets}
\label{app:datasets}

\subsection{Dataset Sources and Statistics}
\label{app:dataset-stats}

\begin{table}[t]
  \centering
  \resizebox{0.98\columnwidth}{!}{%
  \begin{tabular}{lll}
  \toprule
  \textbf{Dataset} & \textbf{Source} & \textbf{Prediction Task} \\
  \midrule
  GitHub Repo  & GH Archive          & Repository event \\
  GitHub User  & GH Archive          & User action \\
  Amazon       & Amazon Reviews'23   & Product category \\
  Earthquake   & USGS API            & Tectonic zone \\
  GDELT        & GDELT 2.0 Database  & Diplomatic action \\
  NBA          & NBA Stats API       & Play type \\
  Wikipedia    & MediaWiki API       & Edit type \\
  \bottomrule
  \end{tabular}}
  \caption{Dataset sources and prediction tasks in \textsc{DescribeEvent}.}
  \label{tab:dataset-overview}
\end{table}

Table~\ref{tab:dataset-overview} lists the seven datasets in \textsc{DescribeEvent} over six domains, together with their sources of data and prediction task.

For the details of their sources, we access each through its public interface and use it within their terms. Specifically, GitHub Repo and GitHub User come from the GH Archive under CC BY-4.0 license; Amazon Reviews from the McAuley Lab's Amazon Reviews 2023 release~\cite{hou2024bridging} for academic research; Earthquake records from the USGS Earthquake Catalog API, a U.S. public-domain federal source; GDELT events from the GDELT 2.0 Event Database, free for academic and non-commercial use with attribution; NBA play-by-play from the public NBA Stats endpoints available for non-commercial use; and Wikipedia edits from the MediaWiki Revisions API, where article content is licensed under CC BY-SA 4.0.

\begin{table*}[!t]
  \centering
  \resizebox{\textwidth}{!}{%
  \begin{tabular}{lcccccccc}
  \toprule
  \textbf{Dataset Statistics} & \textbf{GitHub User} & \textbf{GitHub Repo} & \textbf{Wikipedia} & \textbf{NBA} & \textbf{Earthquake} & \textbf{GDELT} & \textbf{Amazon} & \textbf{Total} \\
  \midrule
  Number of Sequences & 382 & 373 & 276 & 286 & 215 & 179 & 229 & 1,940 \\
  Number of Events & 33,856 & 25,576 & 22,063 & 29,102 & 10,510 & 13,779 & 13,299 & 148,185 \\
  \hdashline
  Number of Event Types & 8 & 8 & 6 & 9 & 8 & 7 & 10 & - \\
  Time Unit & Hours & Hours & Hours & Minutes & Days & Days & Weeks & - \\
  Avg Sequence Length & 88.6 & 68.6 & 79.9 & 101.8 & 48.9 & 77.0 & 58.1 & - \\
  Avg Description Words & 62.3 & 59.9 & 20.8 & 23.8 & 42.9 & 8.6 & 39.4 & -\\
  \bottomrule
  \end{tabular}}
  \caption{Dataset statistics for \textsc{DescribeEvent}.}
  \label{tab:dataset-stats}
\end{table*}

Table~\ref{tab:dataset-stats} reports statistics over all datasets of \textsc{DescribeEvent} after our curation, which contains 1{,}940 sequences and 148{,}185 events in total, with six to ten event types per dataset and average sequence lengths from 48.9 to 101.8 events. Time units range from minutes for NBA plays to weeks for Amazon purchases, and average description length ranges from 8.6 words for GDELT to 62.3 words for GitHub User. For Amazon, the full label space contains 40 product categories. However, each sequence only draws from a single parent category. Therefore in our evaluation, a model is exposed to at most 10 event types per sequence.

\subsection{Semantic Label Space}
\label{app:label-space}

\begin{table*}[!t]
  \centering
  \small
  \begin{tabular}{p{0.09\textwidth}c p{0.78\textwidth}}
  \toprule
  \textbf{Dataset} & $|\mathcal{Y}|$ & \textbf{Semantic Label Space} \\
  \midrule
  Amazon & 10 & \textbf{Beauty \& Personal Care} covers foot/hand/nail care, fragrance, hair care, makeup, men's grooming, personal care, salon \& spa, shave \& hair removal, skin care, and tools \& accessories. \newline \textbf{Books} covers arts \& photography, biographies \& memoirs, children's books, Christian books \& bibles, crafts/hobbies/home, history, literature \& fiction, mystery/thriller/suspense, politics \& social sciences, and teen \& young adult. \newline \textbf{Electronics} covers accessories \& supplies, camera \& photo, car \& vehicle electronics, computers \& accessories, GPS finders \& accessories, headphones/earbuds/accessories, home audio, portable audio \& video, security \& surveillance, and television \& video. \newline \textbf{Home \& Kitchen} covers bath, bedding, event \& party supplies, furniture, heating/cooling/air quality, home d\'{e}cor, kitchen \& dining, seasonal d\'{e}cor, storage \& organization, and wall art. \\
  \addlinespace
  Earthquake & 8 & North America; Central and South America; Northwest Pacific; Southeast Asia and Oceania; Southwest Pacific; Mediterranean and Middle East; Central and South Asia; Atlantic and Africa. \\
  \addlinespace
  GDELT & 7 & public statement; consultation; diplomatic engagement; material cooperation; disapproval; coercion; armed conflict. \\
  \addlinespace
  NBA & 9 & made 2-point; made 3-point; missed field goal; free throw; turnover; foul; rebound; steal or block; timeout. \\
  \addlinespace
  Wikipedia & 6 & revert; reference; content addition; content removal; copy edit; minor edit. \\
  \addlinespace
  GitHub & 8 & issue opened; issue closed; PR opened; PR merged; push; release; PR reviewed; comment. \\
  \bottomrule
  \end{tabular}
  \caption{Label space \texttt{type\_event} for each dataset. $|\mathcal{Y}|$ denotes the number of semantic event labels.}
  \label{tab:label-space}
\end{table*}

Table~\ref{tab:label-space} lists the semantic event types for each dataset, together with the value of $|\mathcal{Y}|$, which denotes the per-sequence label space, namely the number of distinct labels a model can encounter within a single sequence.

For most datasets, $|\mathcal{Y}|$ matches the dataset-wide label space since every sequence draws from the same set of types. But the Amazon dataset is an exception, because each Amazon sequence is anchored to a single parent category, and a model therefore sees only the 10 sub-categories within that category rather than the full 40. We set $|\mathcal{Y}|=10$ accordingly.

\subsection{Sequence Filters}
\label{app:filter-thresholds}

\begin{table*}[!t]
  \centering
  \small
  \begin{tabular}{l cccc p{0.42\textwidth}}
  \toprule
  \textbf{Dataset} & \textbf{Dominant \%} & \textbf{Repeat \%} & \textbf{Min types} & \textbf{Length} & \textbf{Domain-specific filters} \\
  \midrule
  Amazon        & 30\% & 35\%         & 3 & 50--80   & Verified-purchase reviews only; user time span 3--60 months; \emph{other}-typed events stripped. \\
  \addlinespace
  Earthquake    & 50\% & -- & 4 & 30--50   & Magnitude at least 5.0. \\
  \addlinespace
  GDELT         & 35\% & 35\%         & 3 & 40--100  & Root events only; at least 2 mentions, with at least 3 required for armed conflict; business, media, and educational actors excluded from armed conflict; same-country pairs excluded; span 30--1200 days; Goldstein swing of at least 12 on at most 15\% of adjacent events. \\
  \addlinespace
  NBA           & 40\% & 30\%         & 4 & 90--110  & All four regulation quarters present; overtime excluded. \\
  \addlinespace
  Wikipedia     & 40\% & 40\%         & 4 & 60--80   & Bot accounts excluded; chunk duration at most 4380 hours which is about six months. \\
  \addlinespace
  GitHub Repo   & 50\% & -- & 5 & 60--80   & Repositories with more than 20\% bot events dropped; at least 150 characters per event on average; 2--5 distinct actors per repository. \\
  \addlinespace
  GitHub User   & 50\% & -- & 5 & 80--100  & Bot accounts excluded; 2--5 distinct repositories per user; at least 150 characters per event on average; adjacent events with the same type and text collapsed. \\
  \bottomrule
  \end{tabular}
  \caption{Filtering thresholds for each dataset. Dominant \% denotes the maximum share of events with the same \texttt{type\_event} within a sequence, and Repeat \% denotes the maximum share of adjacent event pairs with the same type. Min types specifies the minimum number of distinct event types per sequence, and Length gives the allowed sequence-length range.}
  \label{tab:filter-thresholds}
\end{table*}

This appendix details the curation filters introduced in Section~\ref{sec:curation-filters} with the per-dataset threshold values used to produce \textsc{DescribeEvent}.

There are three sequence-level filters apply to every dataset: \textbf{(i)} The dominant-type cap, denoted Dominant\%, bounds the maximum fraction of events drawn from a single \texttt{type\_event}, \textbf{(ii)} The consecutive-repeat cap, denoted Repeat\%, bounds the maximum fraction of adjacent events sharing the same type; and \textbf{(iii)} The minimum type diversity requirement, or Min types, which sets a floor on the number of distinct types per sequence.

Each sequence must also fall within a per-dataset length range and pass a small set of domain-specific quality filters that remove sources of label noise, such as bot accounts on GitHub and Wikipedia, unverified Amazon reviews, and low-magnitude earthquakes. Any sequence that fails one or more of these checks is discarded before the splits are formed. Table~\ref{tab:filter-thresholds} reports those threshold values for each dataset.

\section{Additional Analyses and Results}
\label{app:additional}

\begin{table}[!t]
  \centering
  \resizebox{\columnwidth}{!}{%
  \begin{tabular}{lccc}
  \toprule
  \textbf{Model} & \textbf{GPQA Diamond} & \textbf{HLE} & \textbf{Desc.\ helps?} \\
  \midrule
  Qwen3.5-27B & 85.8\% & 22.2\% & \cmark \\
  Qwen3.5-9B & 80.6\% & 13.3\% & \cmark \\
  Qwen3.5-4B & 77.1\% & 7.8\% & \cmark \\
  GPT-5 Mini & 80.3\% & 14.6\% & \cmark \\
  Grok-4.1 & 85.3\% & 17.6\% & \cmark \\
  Gemini-2.5 & 79.0\% & 11.1\% & \cmark \\
  \cdashline{1-4}
  GPT-4o Mini & 42.6\% & 4.0\% & \xmark \\
  Llama-3.1-8B & 25.9\% & 5.1\% & \xmark \\
  \bottomrule
  \end{tabular}}
  \caption{Reasoning benchmark scores on GPQA and HLE, and whether adding descriptions improves event prediction.}
  \label{tab:gpqa}
\end{table}

\subsection{Reasoning Benchmark Scores}
\label{app:gpqa}

We use GPQA Diamond and HLE scores as references for the models' reasoning capabilities in Section~\ref{sec:reasoning}. Table~\ref{tab:gpqa} lists those scores for each model, along with whether adding descriptions improves event prediction.

We see a clear split on both benchmarks. Every model that benefits from descriptions scores at least $77\%$ on GPQA Diamond and at least $7\%$ on HLE. GPT-4o Mini and Llama-3.1-8B, for which descriptions hurt, sit far below at $43\%$ and $26\%$ on GPQA Diamond and at $4\%$ and $5\%$ on HLE.

Interestingly, since Qwen3.5-4B is the smallest model in our evaluation but it sits in the helper group with $77.1\%$ on GPQA Diamond, whereas Llama-3.1-8B has twice as many parameters and sits in the non-helper group at $25.9\%$. Together with Figure~\ref{fig:desc-effect} in Section~\ref{sec:reasoning}, these findings support the claim that description utilization tracks general reasoning capability rather than model size.

\subsection{Average Performance Across Datasets}
\label{app:avg-performance}

\begin{table*}[!t]
  \centering
  \small
  \begin{minipage}[t]{0.32\textwidth}
    \centering
    \begin{tabular}{lcc}
    \toprule
    \textbf{Model} & \textbf{F1}$\uparrow$ & \textbf{MAE}$\downarrow$ \\
    \midrule
    \rowcolor{gray!15} \multicolumn{3}{c}{\textit{Semantic Event Type}} \\
    \emph{Qwen3.5-4B} & 0.178 & 5.866 \\
    \emph{Qwen3.5-9B} & 0.221 & 5.684 \\
    \emph{Qwen3.5-27B} & 0.248 & 5.455 \\
    \emph{GPT-5 Mini} & 0.248 & 5.366 \\
    \emph{Grok-4.1} & 0.253 & 5.692 \\
    \emph{Gemini-2.5} & 0.261 & 5.671 \\
    \midrule
    \rowcolor{gray!15} \multicolumn{3}{c}{\textit{Semantic \textbf{w/ Description}}} \\
    \emph{Qwen3.5-4B} & 0.189 & 5.836 \\
    \emph{Qwen3.5-9B} & 0.226 & 5.786 \\
    \emph{Qwen3.5-27B} & 0.266 & 5.389 \\
    \emph{GPT-5 Mini} & 0.259 & 5.378 \\
    \emph{Grok-4.1} & \textbf{0.272} & 5.383 \\
    \emph{Gemini-2.5} & \textbf{0.272} & \textbf{5.350} \\
    \bottomrule
    \end{tabular}
  \end{minipage}%
  \hfill
  \begin{minipage}[t]{0.32\textwidth}
    \centering
    \begin{tabular}{lcc}
    \toprule
    \textbf{Model} & \textbf{F1}$\uparrow$ & \textbf{MAE}$\downarrow$ \\
    \midrule
    \rowcolor{gray!15} \multicolumn{3}{c}{\textit{Symbolic Event Type}} \\
    \emph{Qwen3.5-4B} & 0.184 & 6.463 \\
    \emph{Qwen3.5-9B} & 0.224 & 5.811 \\
    \emph{Qwen3.5-27B} & 0.260 & 5.675 \\
    \emph{GPT-5 Mini} & 0.244 & 5.401 \\
    \emph{Grok-4.1} & 0.239 & 5.821 \\
    \emph{Gemini-2.5} & 0.269 & 5.775 \\
    \midrule
    \rowcolor{gray!15} \multicolumn{3}{c}{\textit{Symbolic \textbf{w/ Description}}} \\
    \emph{Qwen3.5-4B} & 0.207 & 5.673 \\
    \emph{Qwen3.5-9B} & 0.250 & 5.704 \\
    \emph{Qwen3.5-27B} & \textbf{0.290} & \textbf{5.248} \\
    \emph{GPT-5 Mini} & 0.268 & 5.285 \\
    \emph{Grok-4.1} & 0.280 & 5.366 \\
    \emph{Gemini-2.5} & 0.289 & 5.429 \\
    \bottomrule
    \end{tabular}
  \end{minipage}%
  \hfill
  \begin{minipage}[t]{0.32\textwidth}
    \centering
    \begin{tabular}{lcc}
    \toprule
    \textbf{Model} & \textbf{F1}$\uparrow$ & \textbf{MAE}$\downarrow$ \\
    \midrule
    \rowcolor{gray!15} \multicolumn{3}{c}{\textit{non-LLM Baselines}} \\
    \emph{KNN} & \textbf{0.231} & 6.413 \\
    \emph{ARIMA} & 0.146 & 6.806 \\
    \emph{Hawkes} & 0.069 & 19.435 \\
    \emph{NHP} & 0.116 & 5.644 \\
    \emph{RMTPP} & 0.182 & 6.300 \\
    \emph{MLP} & 0.184 & 6.568 \\
    \midrule
    \rowcolor{gray!15} \multicolumn{3}{c}{\textit{non-LLM Baselines \textbf{w/ Description}}} \\
    \emph{NHP} & 0.118 & \textbf{5.461} \\
    \emph{RMTPP} & 0.191 & 6.036 \\
    \emph{MLP} & 0.160 & 6.447 \\
    \bottomrule
    \end{tabular}
  \end{minipage}
  \caption{Dataset-averaged macro F1 and MAE for every row of Tables~\ref{tab:main results} and~\ref{tab:tpp-baselines}, organized into Semantic, Symbolic, and non-LLM groups with their description-aware variants. The best Avg F1 and Avg MAE within each group are in \textbf{bold}.}
  \label{tab:main-avg}
\end{table*}

Table~\ref{tab:main-avg} reports the dataset-averaged macro F1 and MAE for every row of Tables~\ref{tab:main results} and~\ref{tab:tpp-baselines}, summarizing the numbers that underlie Figure~\ref{fig:summary} across the datasets. Notably, Qwen3.5-27B with symbolic labels and descriptions attains the highest Avg F1 of $0.290$ and the lowest Avg MAE of $5.248$ overall, and the four LLM settings ($2 \times 2$ design that crosses description inclusion and label setups) rank in the same order as the finding in our main experiment, with symbolic labels plus descriptions on top and semantic labels alone at the bottom. Adding descriptions raises the best per-group F1 from $0.261$ to $0.272$ under semantic labels and from $0.269$ to $0.290$ under symbolic labels, with the larger gain on the symbolic side, consistent with the semantic bias account. The non-LLM baselines sit well below the LLM rows on F1, with KNN strongest at $0.231$, and pairing NHP, RMTPP, and MLP with descriptions yields a small MAE improvement without closing the F1 gap.

\subsection{Detailed Results on Smaller Models}
\label{app:small-models}

\begin{table*}[!t]
  \centering
  \resizebox{\textwidth}{!}{%
  \begin{tabular}{lcccccccccccccc}
  \toprule
  & \multicolumn{2}{c}{\textbf{GitHub Repo}} & \multicolumn{2}{c}{\textbf{GitHub User}} & \multicolumn{2}{c}{\textbf{Amazon}} &
\multicolumn{2}{c}{\textbf{Earthquake}} & \multicolumn{2}{c}{\textbf{GDELT}} & \multicolumn{2}{c}{\textbf{NBA}} &
  \multicolumn{2}{c}{\textbf{Wikipedia}} \\
  \cmidrule(lr){2-3} \cmidrule(lr){4-5} \cmidrule(lr){6-7} \cmidrule(lr){8-9} \cmidrule(lr){10-11} \cmidrule(lr){12-13} \cmidrule(lr){14-15}
  \textbf{Setting} & F1$\uparrow$ & MAE$\downarrow$ & F1$\uparrow$ & MAE$\downarrow$ & F1$\uparrow$ & MAE$\downarrow$ & F1$\uparrow$ & MAE$\downarrow$ &
F1$\uparrow$ & MAE$\downarrow$ & F1$\uparrow$ & MAE$\downarrow$ & F1$\uparrow$ & MAE$\downarrow$ \\
  \midrule
  \rowcolor{gray!15} \multicolumn{15}{c}{\textit{GPT-4o Mini}} \\
  Symbolic & 0.216 & 2.152 & 0.243 & 1.421 & 0.163 & 5.711 & 0.219 & 0.348 & 0.175 & 10.222 & 0.177 & 0.115 & 0.280 & 16.603 \\
  \quad + Description & 0.199 & 2.276 & 0.183 & 1.518 & 0.187 & 5.780 & 0.202 & 0.391 & 0.162 & 10.139 & 0.088 & 0.107 & 0.255 & 16.701 \\
  \cdashline{1-15}
  Semantic & 0.206 & 2.177 & 0.197 & 1.452 & 0.121 & 5.781 & 0.206 & 0.355 & 0.135 & 10.065 & 0.202 & 0.089 & 0.232 & 16.621 \\
  \quad + Description & 0.218 & 2.325 & 0.171 & 1.563 & 0.131 & 5.849 & 0.199 & 0.371 & 0.155 & 10.179 & 0.163 & 0.104 & 0.249 & 16.639 \\
  \midrule
  \rowcolor{gray!15} \multicolumn{15}{c}{\textit{Llama-3.1-8B}} \\
  Symbolic & 0.224 & 3.958 & 0.216 & 3.084 & 0.141 & 7.251 & 0.190 & 1.294 & 0.175 & 12.454 & 0.154 & 11.877 & 0.237 & 24.908 \\
  \quad + Description & 0.161 & 3.809 & 0.173 & 2.978 & 0.152 & 6.570 & 0.178 & 0.857 & 0.151 & 12.978 & 0.118 & 0.242 & 0.231 & 19.344 \\
  \cdashline{1-15}
  Semantic & 0.187 & 3.983 & 0.182 & 2.712 & 0.139 & 7.717 & 0.165 & 4.639 & 0.153 & 12.413 & 0.117 & 0.171 & 0.225 & 31.863 \\
  \quad + Description & 0.139 & 4.354 & 0.125 & 2.616 & 0.141 & 6.699 & 0.182 & 0.880 & 0.155 & 15.325 & 0.110 & 0.118 & 0.195 & 19.094 \\
  \bottomrule
  \end{tabular}}
  \caption{Performance of smaller models across datasets. Adding descriptions does not consistently help, suggesting that sufficient model capability is needed to leverage descriptive information.}
  \label{tab:small-models}
\end{table*}

This appendix expands the results in Section~\ref{sec:reasoning} with two smaller models, GPT-4o Mini and Llama-3.1-8B, run under the same four-setting protocol as the capable LLMs in Table~\ref{tab:main results}. Table~\ref{tab:small-models} reports macro F1 and MAE for all four settings on the seven datasets in \textsc{DescribeEvent}.

Adding descriptions does not improve overall macro F1 for either model. Specifically, on GPT-4o Mini, descriptions reduce overall F1 by 0.028 under symbolic labels and by 0.002 under semantic labels. On Llama-3.1-8B, the drop is 0.025 under symbolic labels and 0.018 under semantic labels. The drop is therefore larger under symbolic than under semantic labels for both models, inverting the pattern seen in the capable LLMs in our main Table~\ref{tab:main results}, where symbolic labels paired with descriptions is the strongest setting.

\subsection{Detailed Analysis about Rare Events}
\label{app:rare-event-all}

\begin{figure*}[!t]
  \centering
  \includegraphics[width=\textwidth]{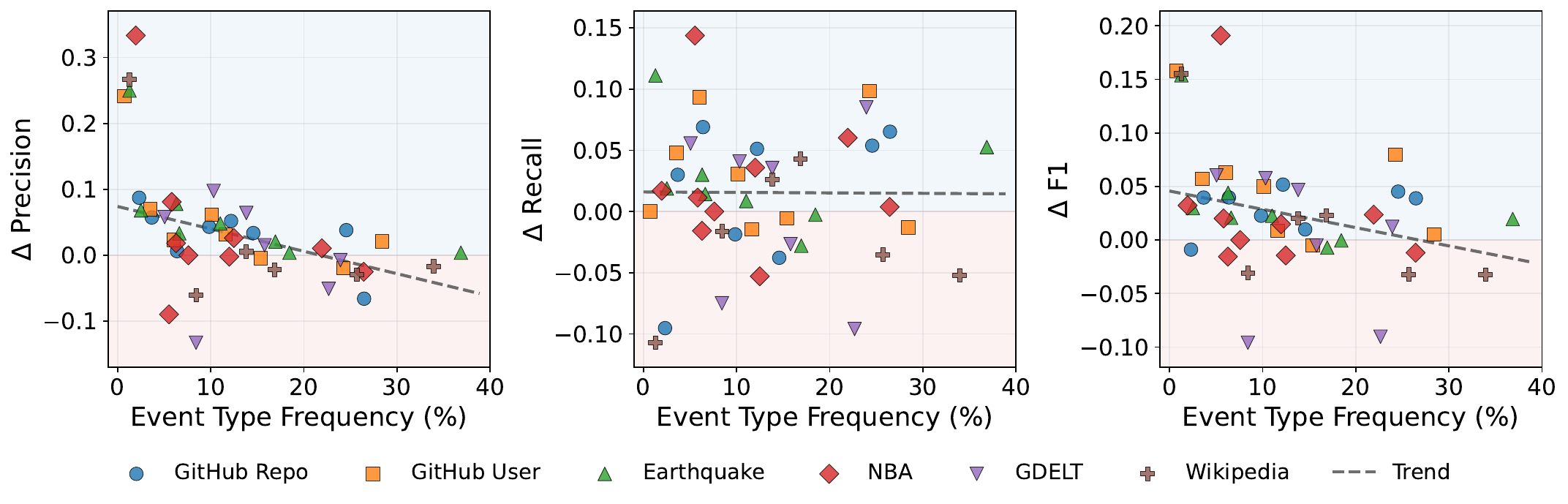}
  \caption{Per-type gain from Symbolic over Semantic labels, both with descriptions, versus event type frequency, for precision, recall, and F1.}
  \label{fig:rare-event-all}
\end{figure*}

Figure~\ref{fig:rare-event} in Section~\ref{sec:rare-event} only provide plots for the precision panel of the per-event gain analysis. In this appendix, we shows all three panels in Figure~\ref{fig:rare-event-all}, with precision, recall, and F1 from left to right. Each point is one event type from the six datasets, plotted at its within-domain frequency against the symbolic-over-semantic performance gain. Precision and F1 both rise as frequency decreases, with correlations of $r=-0.51$ and $r=-0.40$, while recall stays flat at $r=-0.01$. This matches our semantic bias intuition. Symbolic labels remove the bias-driven false positives that dilute precision on rare events, while the description that drives recall is shared by both label settings.

\section{LLM Inference Configuration}
\label{app:inference}

We serve the open-weight Qwen3.5 models locally with vLLM~\cite{kwon2023vllm}, and access the closed-weight models GPT-5 Mini, Grok-4.1 Fast, Gemini-2.5 Flash, and GPT-4o Mini through the OpenRouter API. For each model we keep the provider's default sampling temperature and do not tune any decoding hyperparameters, so the comparison reflects out-of-the-box behavior. For models that support a deep thinking mode, we disable it because extended traces tend to inflate intermediate computation on simple inputs without improving accuracy~\cite{chen2025overthinking}, and we observed the same degradation in our preliminary experiments. We instead instruct the model to write a concise chain of thought of two to three sentences before emitting the final \{\texttt{event\_type}, \texttt{time}\} answer, as shown in the system message in Figure~\ref{fig:prompts}.

For a sequence of length $N$, we set the initial history length to $n_0 = \lfloor \rho N \rfloor$, where the fraction $\rho \in (0, 1)$ is chosen per dataset. Specifically, for the datasets with shortest averaged sequence length shown in in Table~\ref{tab:dataset-stats}, i.e., Earthquake and Amazon, we set $\rho = 0.8$, and for the rest, we set $\rho = 0.9$. This yields roughly eight prediction steps per sequence on average across all datasets.

\section{Prompt Templates}
\label{app:prompts}

\begin{figure*}[t]
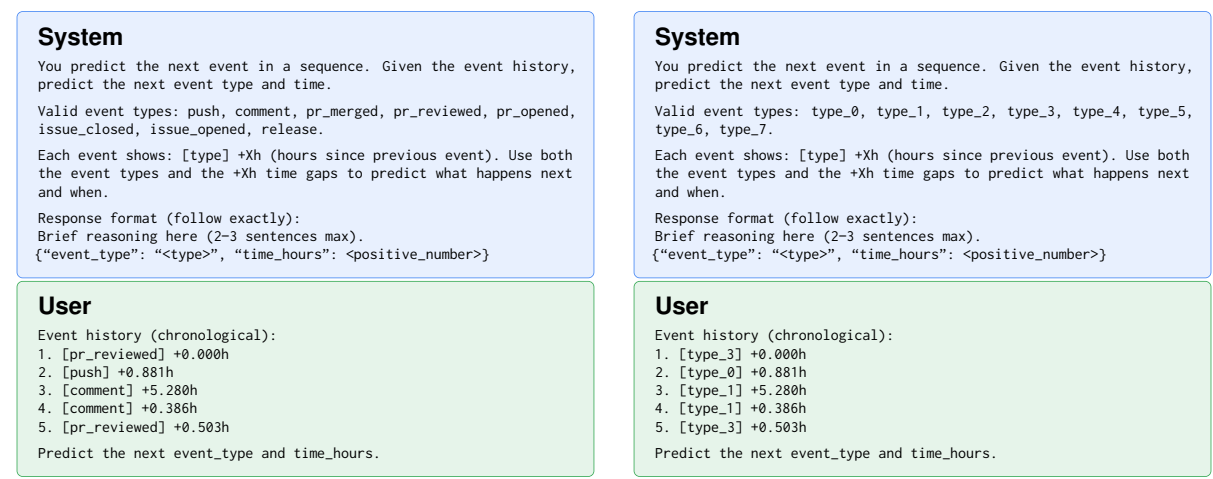
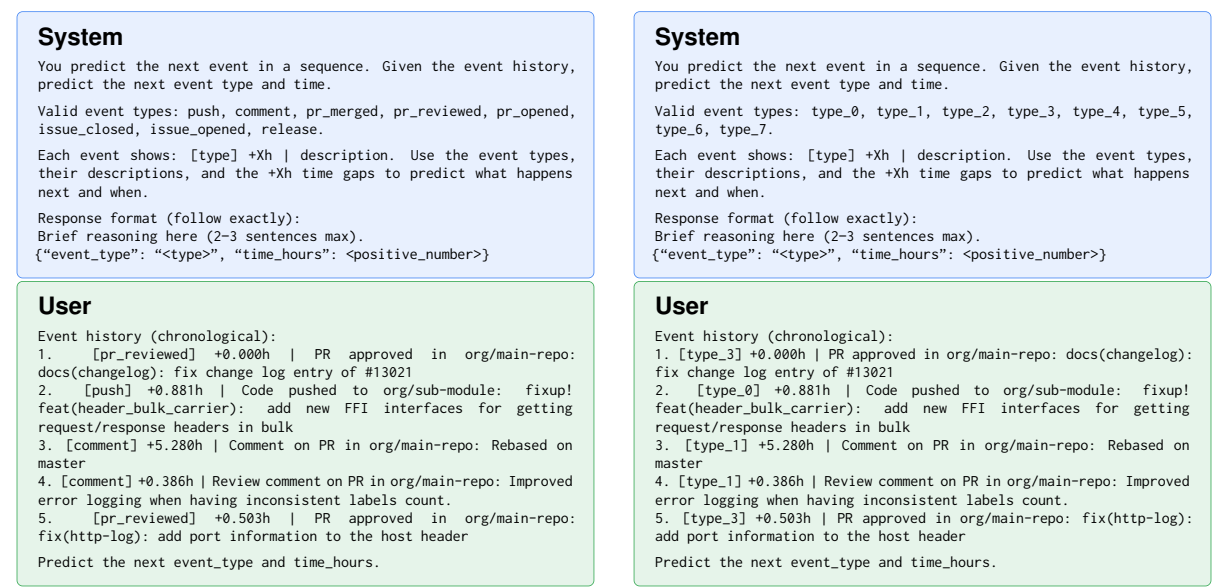

\centering
\scriptsize
\begin{subfigure}[!t]{0.49\textwidth}
\begin{promptbox}
\begin{sysmsg}
You predict the next event in a sequence. Given the event history, predict the next event type and time.\par\smallskip
Valid event types: push, comment, pr\_merged, pr\_reviewed, pr\_opened, issue\_closed, issue\_opened, release.\par\smallskip
Each event shows: [type] +Xh (hours since previous event). Use both the event types and the +Xh time gaps to predict what happens next and when.\par\smallskip
Response format (follow exactly):\par
Brief reasoning here (2--3 sentences max).\par
\{``event\_type'': ``<type>'', ``time\_hours'': <positive\_number>\}
\end{sysmsg}
\smallskip
\begin{usrmsg}
Event history (chronological):\par
1. [pr\_reviewed] +0.000h\par
2. [push] +0.881h\par
3. [comment] +5.280h\par
4. [comment] +0.386h\par
5. [pr\_reviewed] +0.503h\par\smallskip
Predict the next event\_type and time\_hours.
\end{usrmsg}
\end{promptbox}
\caption{\textbf{Semantic} setting. Human-readable names, no descriptions.}
\label{fig:prompt-semantic}
\end{subfigure}\hfill
\begin{subfigure}[!t]{0.49\textwidth}
\begin{promptbox}
\begin{sysmsg}
You predict the next event in a sequence. Given the event history, predict the next event type and time.\par\smallskip
Valid event types: type\_0, type\_1, type\_2, type\_3, type\_4, type\_5, type\_6, type\_7.\par\smallskip
Each event shows: [type] +Xh (hours since previous event). Use both the event types and the +Xh time gaps to predict what happens next and when.\par\smallskip
Response format (follow exactly):\par
Brief reasoning here (2--3 sentences max).\par
\{``event\_type'': ``<type>'', ``time\_hours'': <positive\_number>\}
\end{sysmsg}
\smallskip
\begin{usrmsg}
Event history (chronological):\par
1. [type\_3] +0.000h\par
2. [type\_0] +0.881h\par
3. [type\_1] +5.280h\par
4. [type\_1] +0.386h\par
5. [type\_3] +0.503h\par\smallskip
Predict the next event\_type and time\_hours.
\end{usrmsg}
\end{promptbox}
\caption{\textbf{Symbolic} setting. Anonymous labels, no descriptions.}
\label{fig:prompt-symbolic}
\end{subfigure}

\vspace{2em}

\begin{subfigure}[!t]{0.49\textwidth}
\begin{promptbox}
\begin{sysmsg}
You predict the next event in a sequence. Given the event history, predict the next event type and time.\par\smallskip
Valid event types: push, comment, pr\_merged, pr\_reviewed, pr\_opened, issue\_closed, issue\_opened, release.\par\smallskip
Each event shows: [type] +Xh | description. Use the event types, their descriptions, and the +Xh time gaps to predict what happens next and when.\par\smallskip
Response format (follow exactly):\par
Brief reasoning here (2--3 sentences max).\par
\{``event\_type'': ``<type>'', ``time\_hours'': <positive\_number>\}
\end{sysmsg}
\smallskip
\begin{usrmsg}
Event history (chronological):\par
1. [pr\_reviewed] +0.000h | PR approved in org/main-repo: docs(changelog): fix change log entry of \#13021\par
2. [push] +0.881h | Code pushed to org/sub-module: fixup! feat(header\_bulk\_carrier): add new FFI interfaces for getting request/response headers in bulk\par
3. [comment] +5.280h | Comment on PR in org/main-repo: Rebased on master\par
4. [comment] +0.386h | Review comment on PR in org/main-repo: Improved error logging when having inconsistent labels count.\par
5. [pr\_reviewed] +0.503h | PR approved in org/main-repo: fix(http-log): add port information to the host header\par\smallskip
Predict the next event\_type and time\_hours.
\end{usrmsg}
\end{promptbox}
\caption{\textbf{Semantic + Description} setting. Human-readable names with descriptions.}
\label{fig:prompt-semantic-desc}
\end{subfigure}\hfill
\begin{subfigure}[!t]{0.49\textwidth}
\begin{promptbox}
\begin{sysmsg}
You predict the next event in a sequence. Given the event history, predict the next event type and time.\par\smallskip
Valid event types: type\_0, type\_1, type\_2, type\_3, type\_4, type\_5, type\_6, type\_7.\par\smallskip
Each event shows: [type] +Xh | description. Use the event types, their descriptions, and the +Xh time gaps to predict what happens next and when.\par\smallskip
Response format (follow exactly):\par
Brief reasoning here (2--3 sentences max).\par
\{``event\_type'': ``<type>'', ``time\_hours'': <positive\_number>\}
\end{sysmsg}
\smallskip
\begin{usrmsg}
Event history (chronological):\par
1. [type\_3] +0.000h | PR approved in org/main-repo: docs(changelog): fix change log entry of \#13021\par
2. [type\_0] +0.881h | Code pushed to org/sub-module: fixup! feat(header\_bulk\_carrier): add new FFI interfaces for getting request/response headers in bulk\par
3. [type\_1] +5.280h | Comment on PR in org/main-repo: Rebased on master\par
4. [type\_1] +0.386h | Review comment on PR in org/main-repo: Improved error logging when having inconsistent labels count.\par
5. [type\_3] +0.503h | PR approved in org/main-repo: fix(http-log): add port information to the host header\par\smallskip
Predict the next event\_type and time\_hours.
\end{usrmsg}
\end{promptbox}
\caption{\textbf{Symbolic + Description} setting. Anonymous labels with descriptions.}
\label{fig:prompt-symbolic-desc}
\end{subfigure}
\caption{Prompt templates for the four evaluation settings, illustrated on a GitHub User event sequence.}
\label{fig:prompts}
\end{figure*}

We illustrate the four evaluation settings using a GitHub User event sequence as a running example, shown in Figure~\ref{fig:prompts}. The settings differ along two axes, whether event types use \emph{semantic} labels such as \texttt{push} and \texttt{comment} or \emph{symbolic} labels such as \texttt{type\_0} and \texttt{type\_1}, and whether natural-language descriptions are included.

The system message tells the model that it must provide a brief reasoning with two to three sentences and predict the next event type and its time. It also lists the valid event types for the sequence, and fixes the response format so that we can parse a single event type and a positive number of hours. The user message then serializes the history as a chronological list of past events, with each event written as \texttt{[type] +Xh}, where \texttt{+Xh} is the time gap from the previous event. The description variants append a short natural-language sentence after each event, drawn from \textsc{DescribeEvent}.

\end{document}